\documentclass[letterpaper]{article} 
\usepackage{aaai2026}  
\usepackage{times}  
\usepackage{helvet}  
\usepackage{courier}  
\usepackage[hyphens]{url}  
\usepackage{graphicx} 
\usepackage{natbib}  
\usepackage{caption} 
\nocopyright
\usepackage{graphicx}     
\usepackage{adjustbox}    

\usepackage[table]{xcolor}
\usepackage{algorithm}
\usepackage{algorithmic}
\usepackage{bbding}
\usepackage{tabularx}
\usepackage{arydshln}  
\usepackage{booktabs}  
\usepackage{array}
\usepackage{multirow}  
\usepackage{amsmath}
\usepackage{amssymb}
\usepackage{subcaption}
\usepackage[utf8]{inputenc} 
\usepackage{url}            
\usepackage{amsfonts}       
\usepackage{nicefrac}       
\usepackage{microtype}      
\usepackage{xcolor}         
\usepackage{enumitem}

\definecolor{emotionHappy}{RGB}{255, 249, 196}    
\definecolor{emotionSad}{RGB}{187, 222, 251}      
\definecolor{emotionNeutral}{RGB}{224, 224, 224}  

\usepackage{newfloat}
\usepackage{listings}
\DeclareCaptionStyle{ruled}{labelfont=normalfont,labelsep=colon,strut=off} 
\floatstyle{ruled}
\newfloat{listing}{tb}{lst}{}
\floatname{listing}{Listing}
\title{
EmoTransCap: Dataset and Pipeline for Emotion Transition-Aware Speech Captioning in Discourses}

\author{
Shuhao Xu\textsuperscript{\rm 1},
Yifan Hu\textsuperscript{\rm 1},
Jingjing Wu\textsuperscript{\rm 1}, \\
Zhihao Du,
Zheng Lian\textsuperscript{\rm 2},
Rui Liu\textsuperscript{\rm 1}\thanks{Corresponding Author.}
}

\affiliations{
\textsuperscript{\rm 1}Inner Mongolian University, China\\
\textsuperscript{\rm 2}State Key Laboratory of Autonomous Intelligent Unmanned Systems, Tongji University\\
\{0231122657, 22309013, 0221121690\}@mail.imu.edu.cn,
duzhihao.china@gmail.com, imucslr@imu.edu.cn
}

\usepackage{bibentry}

\begin{document}
\maketitle
\begin{abstract}
Emotion perception and adaptive expression are fundamental capabilities in human-agent interaction. While recent advances in speech emotion captioning (SEC) have enhanced emotion perception and expression through fine-grained emotional modeling, existing systems remain limited to static, single-emotion characterization within isolated sentences, neglecting the inherently dynamic emotional transitions that occur at the discourse level. To address this critical gap, we propose a novel paradigm that integrates temporal emotion dynamics with discourse-level speech description, termed \textit{Emotion Transition-Aware Speech Captioning (EmoTransCap)}:
\textbf{1) Data.} 
To construct a dataset rich in emotion transitions while enabling scalable expansion in a time- and labor-efficient manner, we design an automated pipeline for dataset creation. This is the first large-scale dataset explicitly designed to capture discourse-level emotion transitions.
\textbf{2) Annotation.} 
To generate semantically rich descriptions, we propose a method that incorporates acoustic attributes and temporal cues from discourse-level speech. Notably, our temporal cues recognition relies on an innovative \textit{Multi-Task Emotion Transition Recognition (MTETR)} that performs joint emotion transition detection and diarization. At last, leveraging the strong semantic analysis capabilities of LLMs, we produce two annotation versions: descriptive and instruction-oriented.
\textbf{3) Performance.}
The above data and annotations offer a valuable resource for advancing both emotion perception and emotional expressiveness. For emotion perception, this dataset enables the generation of speech captions that capture emotional transitions, facilitating temporal-dynamic and fine-grained emotion understanding. For emotion expression, we introduce the first controllable, transition-aware emotional speech synthesis system at the discourse level, significantly enhancing anthropomorphic emotional expressiveness. This work establishes a new foundation for developing emotionally intelligent conversational agents. 
\end{abstract}

\begin{figure*}[h!]
\centering
\small
\centerline{
\includegraphics[width=1\linewidth]{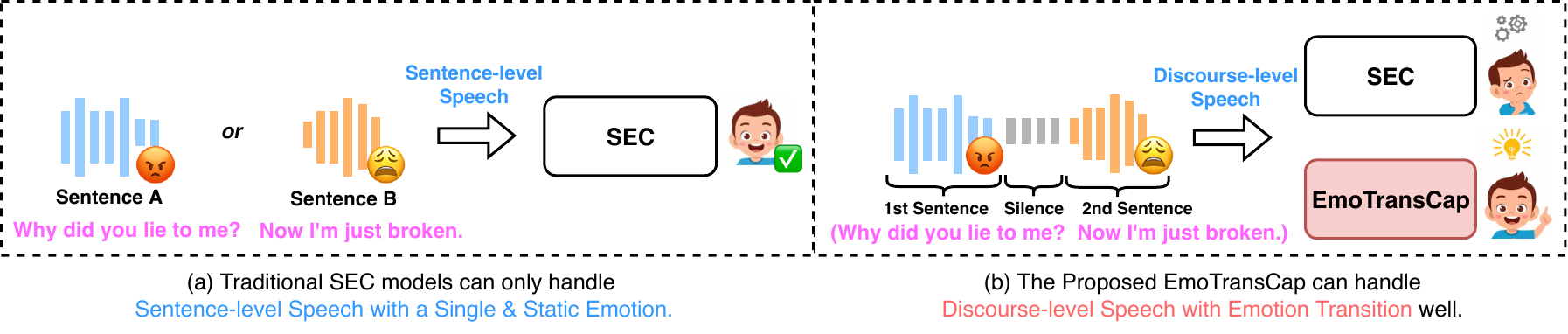}
}
\caption{Illustration of the basic idea of our EmoTransCap, which provides an accurate description of emotion-transition cues at the discourse level. (Take discourse with 1 transition as an example.)}
\label{fig:fig1}
\end{figure*}

\section{Introduction}
Accurate emotional perception and anthropomorphic emotional expression are essential for developing empathetic conversational agents \cite{ball2000emotion}. These two capabilities enable conversational systems to dynamically recognize user affect \cite{poria2019emotion}, contextually adapt emotional responses \cite{zhou2018emotional}, and generate human-like prosody with natural affective rendering \cite{im2022emoq}. They facilitate the application of conversational agents across diverse domains, including education \cite{pekrun2014introduction}, mental health \cite{bendelow2009health}, and finance \cite{ackert2003emotion}.
To capture compound and fine-grained emotional states in human emotions, researchers propose speech emotion captioning (SEC) \cite{xu2023secapspeechemotioncaptioning}, which employs human-like free-form language descriptions to depict emotional states.
Given its advantages, SEC has emerged as a prominent research focus \cite{yan-etal-2024-talk}, with numerous speech-related tasks leveraging SEC to improve the accuracy of emotional perception and the naturalness of emotion expression \cite{jin2024speechcraft,zhu2024unistyle,yamauchi2024stylecap,lian2025affectgpt}.

Although SEC advances empathetic human-agent speech interaction, several critical challenges remain unresolved. \textbf{1) Single and Static Emotional Characterization}: Existing approaches typically assume that a speech sentence contains only a single, static emotional state. However, human emotions are inherently dynamic, often exhibiting emotional transitions from one state to another (e.g., frustration-to-joy progression during conflict resolution) \cite{cowie2003describing}. Accurately describing these dynamic emotional processes is fundamental for modeling user characteristics, predicting emotions, and inducing emotional responses \cite{kuppens2017emotion}. \textbf{2) Isolated Sentence Modeling}: Current methods predominantly analyze speech at the isolated sentence level. However, as previously mentioned, emotional transitions frequently occur across sentences at the discourse level \cite{filipowicz2011understanding}. When processing discourse-level inputs, researchers often first analyze individual sentences and then integrate their results \cite{Gu_Yoo_Ha_2021}, which fails to capture the coherence among utterances \cite{cacioppo1999emotion,okon2013dynamic}. These limitations hinder the development of speech interaction techniques at the discourse level between humans and agents.

To bridge this gap, we propose a novel paradigm that integrates temporal emotion transitions with discourse-level speech description, termed \textit{Emotion Transition-Aware Speech Captioning (EmoTransCap)}. Figure \ref{fig:fig1} compares our approach with previous sentence-level speech captioning methods that rely on single, static emotion labels. To further advance this critical task, we build a dedicated dataset and introduce a novel annotation methodology. \textbf{1) Dataset.}
Due to the lack of speech datasets with rich emotion transitions, we leverage the powerful semantic comprehension capabilities of Gemma-3 \cite{team2025gemma} and the speech generation capabilities of CosyVoice2 \cite{du2024cosyvoice} to automatically construct a bilingual (Chinese and English) discourse-level speech dataset, termed {\textit{EmoTransSpeech-Audio}}. This automated data generation pipeline enables scalable expansion of the dataset in a time- and labor-efficient manner. To the best of our knowledge, this is the first large-scale speech dataset featuring rich emotion transitions.
\textbf{2) Annotation.}
We design an automated pipeline to identify acoustic attributes and temporal cues in discourse-level speech and employ Gemma-3 to generate the final captions, denoted as \textit{EmoTransSpeech-Caption}. Notably, our temporal cue recognition relies on an innovative \textit{Multi-Task Emotion Transition Recognition Model (MTETR)} that performs joint emotion transition detection and diarization. This annotation approach enables the generation of semantically rich descriptions with minimal manual effort.
Finally, our dataset contains both descriptive and instruction-oriented versions, serving as a valuable resource for emotion perception and generation tasks.

Extensive subjective and objective evaluations demonstrate that EmoTransCap surpasses baselines on key metrics, excelling particularly in modeling discourse-level emotional dynamics. Furthermore, leveraging EmoTransCap, we present the first realization of controllable and transition-aware emotional speech synthesis, which achieves nuanced emotion rendering for discourse. These advancements establish new benchmarks for context-aware emotion understanding and human-like expressive synthesis. Our main contributions are summarized as follows:
\textbf{$\bullet$ Task.} 
    This work pioneers emotion transition-aware captioning at the discourse level, representing an advancement beyond existing sentence-level approaches that rely on a single, static emotion. Our work promotes the development of emotionally intelligent human-agent interaction.
    
 \textbf{$\bullet$ Foundation.} 
    We introduce the first large-scale bilingual discourse-level emotional speech dataset, EmoTransSpeech
    \footnote{EmoTransSpeech-Audio and EmoTransSpeech-Caption.}
    , featuring rich emotion transitions. Meanwhile, we design a fully automated pipeline, which can identify emotional transitions within discourse-level utterances and generate semantically rich captions. This resource serves as a critical foundation for advancing accurate emotion perception and anthropomorphic emotional expression.
    
    
 \textbf{$\bullet$ Performance.} 
    Extensive subjective and objective evaluations validate the effectiveness of EmoTransCap. For emotion perception, our method demonstrates superior capability in accurately capturing dynamic emotional transitions within discourses. For emotion expression, we achieve the first discourse-level controllable emotional transition speech synthesis.

 


    


\section{Related Works}
\textbf{Speech Emotional Captioning.}
\textbf{(1) Model.} Traditional speech emotion recognition (SER) categorizes emotions into fixed classes \cite{ekman1992argument}, while speech emotion captioning (SEC) generates natural language descriptions of emotional states \cite{xu2023secapspeechemotioncaptioning}. Recent SEC advancements include SECap \cite{xu2023secapspeechemotioncaptioning} for Language-based emotion understanding \cite{xu2023secapspeechemotioncaptioning}, SpeechCraft for fine-grained stylized descriptions \cite{jin2024speechcraft}, and AlignCap for human preference-aligned outputs \cite{liang2024aligncap}. For expression tasks, methods like InstructTTS \cite{yang2024instructtts}, PromptTTS \cite{leng2023prompttts}, and Salle \cite{ji2023textrolspeechtextstylecontrol}, etc. integrate SEC with controllable synthesis. However, existing approaches focus on static and singular emotional states, isolated utterances and neglect discourse-level dynamics. Our EmoTransCap overcomes these limitations by modeling emotional transitions in discourse. Additionally, we propose dual caption formats (descriptive/instructional) to better support downstream applications.
\textbf{(2) Dataset.}
Speech-caption paired datasets are critical for SEC research. Existing efforts include: PromptSpeech \cite{leng2023prompttts}: Expert-annotated captions for synthesized and real speech. TextrolSpeech \cite{ji2024textrolspeech}: GPT-generated style descriptions scaled via multi-stage prompting. SpeechCraft \cite{jin2024speechcraft}: Combines expert classifiers with LLaMA2 for attribute-rich captions. ZED \cite{wang2023speech}: Provides manually annotated boundaries for emotion segments but focuses on isolated sentences.
While these datasets advance emotion understanding, they lack discourse-level emotional transition annotations, limiting their utility for modeling temporal dynamics across sentences. To address this, we propose a synthetic dataset with fine-grained natural language descriptions of emotion transition in discourse, bridging a key gap in emotionally intelligent agent development.

\begin{figure*}[t]
\small
\centering
\centerline{
\includegraphics[width=1\linewidth]{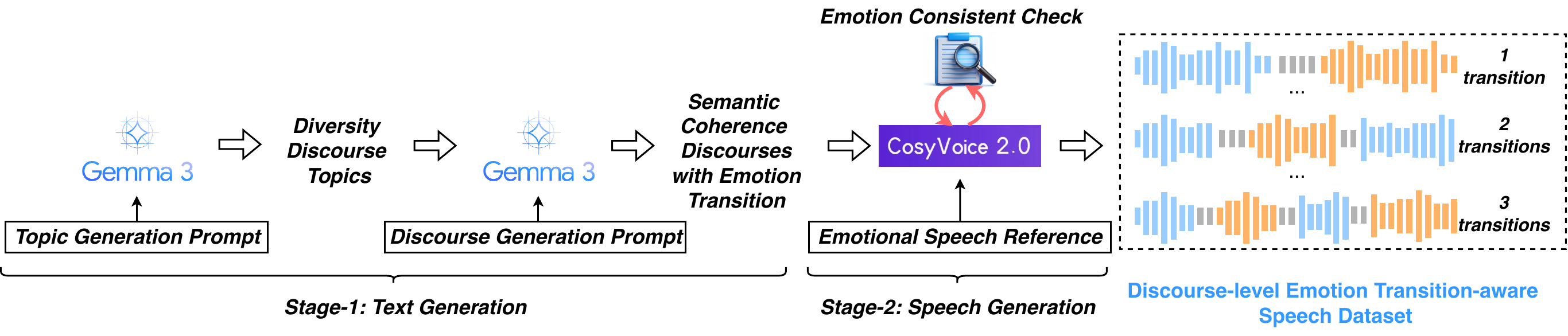}
}
\caption{The pipeline of the discourse-level emotion transition-aware speech dataset construction. Gray speech waveforms represent silent segments.}
\label{fig:fig2}
\end{figure*}

\begin{figure*}[t]
\small
\centering
\centerline{
\includegraphics[width=1\linewidth]{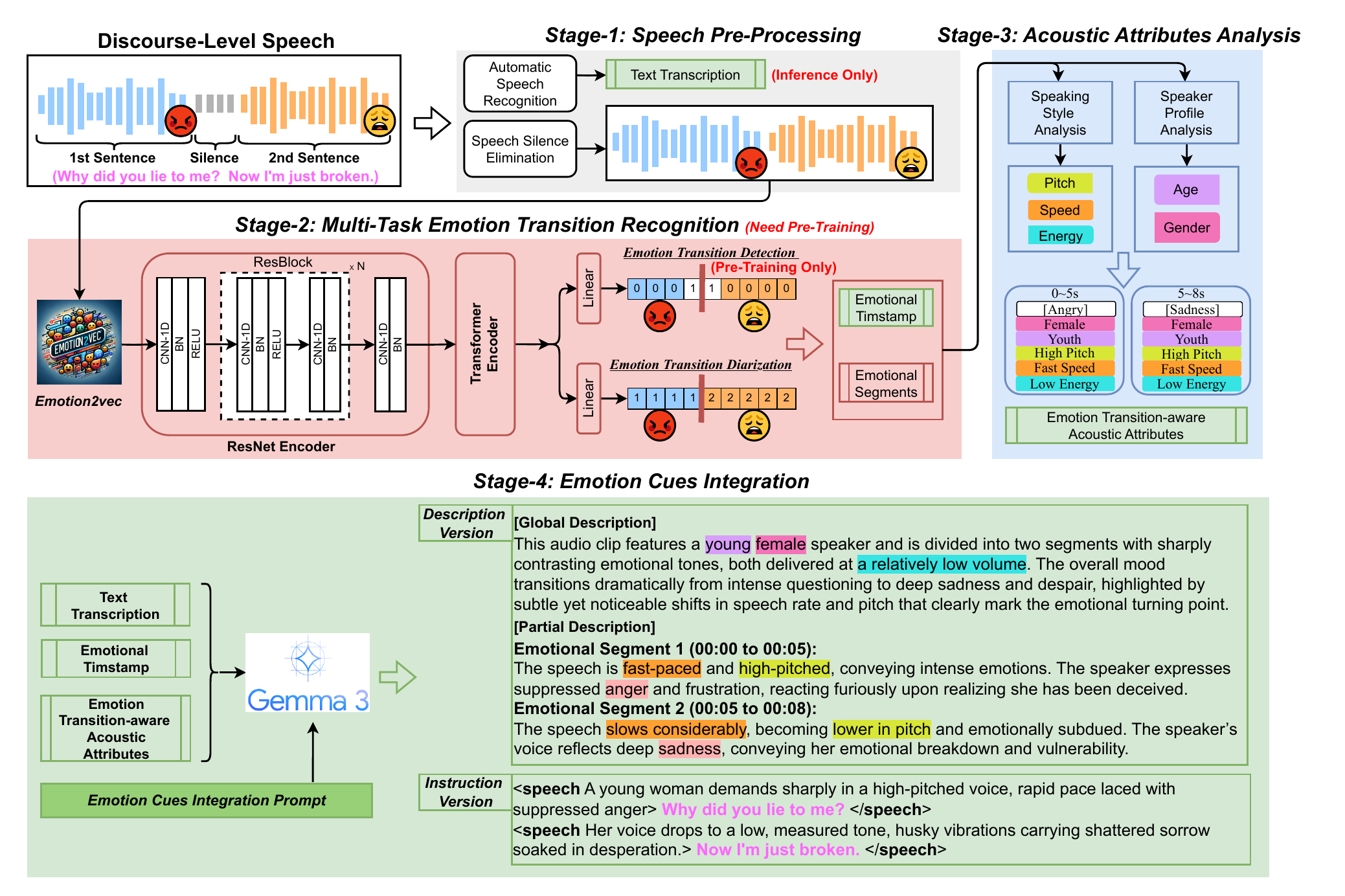}
}
\caption{The overall workflow of EmoTransCap annotation pipeline. (Take discourse with 1 transition as an example.)}
\label{fig:fig3}
\end{figure*}

\textbf{Speech Emotion Diarization.}
In psychology, emotion modeling primarily follows categorical and dimensional approaches \cite{cacioppo1999emotion}. The dimensional framework represents emotions through continuous scales (e.g., arousal, valence) \cite{mehrabian1974approach}, while the categorical approach classifies emotions into discrete classes like Ekman's six basic emotions \cite{ekman1992argument} and Plutchik’s eight primary emotions \cite{plutchik1980general}, favored for its intuitiveness and annotation efficiency. Speech Emotion Diarization (SED) extends traditional categorical emotion recognition by identifying both emotion categories and their temporal boundaries \cite{wang2023speech}, capturing dynamic emotional shifts in speech. Recent studies leverage self-supervised models (e.g., Wav2vec 2.0 \cite{baevski2020wav2vec}, Hubert \cite{hsu2021hubert}, WavLM \cite{chen2022wavlm}) with linear classifiers for frame-level predictions \cite{wang2023speech,tang2024ed}.
Our proposed Multi-Task Emotion Transition Recognition (MTETR) differs from existing SED methods in three key aspects: 1) Replacing standard self-supervised models with Emotion2vec \cite{ma2023emotion2vecselfsupervisedpretrainingspeech}, a specialized emotion representation model; 2) Employing a ResNet-Transformer-BiLSTM architecture to model long-range temporal dependencies; 3) Introducing a multi-task framework that combines emotion transition detection with diarization to enhance contextual learning.
\textbf{Advanced Emotional Modeling.}
As mentioned before, for standard categorical speech emotion recognition, researchers typically represent a character’s emotional state using the most likely label from a predefined taxonomy of basic emotions. However, human emotions are often mixed in certain scenarios \cite{du2014compound}, prompting the use of multiple labels within the basic emotion framework \cite{li2017reliable}. To capture more nuanced emotional expressions beyond these basic categories, researchers introduce open-vocabulary emotion models \cite{lian2025ov,lian2023explainable}, enabling the generation of unrestricted emotional categories and label numbers. Nevertheless, all these approaches remain within the realm of temporal-independent annotations, which assign one or more labels to a sentence without accounting for the temporal dynamics of emotions. 
However, emotions evolve dynamically, even at the discourse level. This phenomenon, known as \textit{emotional transition} \cite{sun2019dynamic}, is well-established in psychology \cite{filipowicz2011understanding} but remains underexplored in computer science. To address this gap, we propose a novel paradigm that integrates temporal emotion dynamics with discourse-level speech description, while also constructing datasets to advance this field. This work promotes the development of more accurate, temporal-dependent emotion representations and their applications in emotion perception and expression.


\section{Discourse-level Emotion Transition-aware Speech Dataset}\label{sec3}

In this paper, we introduce and release a large-scale, bilingual, discourse-level speech dataset with rich emotion transitions. 
As shown in Figure. \ref{fig:fig2}, we propose a two-stage pipeline, with the first stage responsible for generating semantically coherent and expressive discourse-level text and the second stage responsible for synthesizing speech signals for all discourses with expected emotion transitions.

\textbf{Stage-1: Text Generation.} 
\textbf{1) Topic Generation}: To ensure broad coverage of real-world speaking scenarios in our dataset, we design a hierarchical discourse topic selection strategy. First, inspired by existing research on topic classification \cite{ramzan2023advancements}, we identify seven high-level topic categories as primary topics: [Business, Culture, Daily Life, Entertainment, Politics, Science, Sports]. 
Then, leveraging the extensive knowledge base and generative capabilities of Gemma-3, we generate a set of more fine-grained secondary topics for each primary topic. Finally, through manual curation, we select 60 representative secondary topics for each primary topic, resulting in a total of 420 distinct secondary discourse topics.
This hierarchical strategy provides a diverse and comprehensive semantic foundation for subsequent discourse generation.
\textbf{2) Discourse Generation}: Based on the generated topics, we further prompt the Gemma-3 model to generate the semantically coherent discourses, with emotional transitions, in bulk. Note that we construct 420 combinations of emotional transitions by varying both the emotion categories (angry, happy, sad, neutral, and surprised) and the segments of the discourse. 
The number of emotion combination types under different numbers of emotion transitions in discourses: 1 transition: 20 combinations; 2 transitions: 80 combinations; 3 transitions: 320 combinations. 
To enhance the realism and diversity of the generated texts, we incorporate first-person, second-person, and third-person narrative perspectives, which are commonly found in real-world speaking scenarios~\cite{liu2024analysis} and are explicitly considered during the discourse generation process.
Please refer to the Appendix A.1 for details on the \textit{Topic and Discourse Generation Prompts}.


\textbf{Stage-2: Speech Generation.} \textbf{1) Speech Synthesis:} Given that existing TTS models are not yet capable of end-to-end generation of discourse-level emotion transition-aware speech, we adopt a sentence-wise synthesis strategy to generate speech for each discourse. Specifically, we adopt CosyVoice2~\cite{du2024cosyvoice2scalablestreaming}, a powerful zero-shot voice cloning system, to synthesize each sentence of discourse conditioned on the emotional speech reference with the target emotion to produce emotion-consistent and stylistically appropriate speech \footnote{We have also tried to use language description to guide emotional speech generation, but we found that its timbre and emotional stability were not as good as using speech prompt.}. 
Note that the reference emotional speech prompts are selected from the following two commonly used emotional speech datasets, ESD~\cite{zhou2021seenunseenemotionalstyle} and RAVDESS~\cite{livingstone2018ryerson}, from which we curate 20 speakers covering five basic emotions. 
\textbf{2) Emotion Consistent Check}: 
For each synthesized sentence, we adopt the pre-trained emotion2vec \cite{ma2023emotion2vecselfsupervisedpretrainingspeech} as the SER model to conduct the emotion-consistent checking.
When the emotion state label of synthetic speech after emotion recognition is inconsistent with the label defined in Stage-1, it needs to be forcibly re-synthesized until the emotion label is consistent.
This ensures that the intended emotion in all speech is accurately conveyed. 
After generating all individual sentence speech, we sequentially concatenate them based on predefined emotional transition combinations of Stage-1. To improve perceptual continuity, we apply average loudness normalization to each segment prior to concatenation, which reduces inter-segment volume discrepancies and ensures a smoother and more natural emotional progression.

At last, the \textit{EmoTransSpeech-Audio} dataset comprises 144,000 utterances featuring 1 to 3 emotional transitions, synthesized by 10 native English speakers and 10 native Mandarin speakers, with an equal distribution of 5 males and 5 females per language. The total duration of the dataset exceeds 617 hours. 
For more details, please refer to the Appendix A.1.

\section{EmoTransCap: Methodology}
\textbf{Pipeline Overview.}
As shown in Figure. \ref{fig:fig3}, the proposed automatic \textit{EmoTransCap} pipeline operates in the following four stages, that are [Stage-1: Speech Pre-processing]; [Stage-2: Multi-Task Emotion Transition Recognition]; [Stage-3: Acoustic Attributes Analysis]; and [Stage-4: Emotion Cues Integration].
Through these four stages, \textit{EmoTransCap} produces fine-grained, stylistically diverse, and emotion transition-aware natural language captions for discourse-level speech.

\textbf{Stage-1: Speech Pre-Processing.} Since the non-speech fragments in speech will bring redundant information to the speech emotion dynamic analysis \cite{wang2023speech}, the first step of speech pre-processing is to eliminate the silent segments in the input discourse-level speech. After that, we also get the linguistic information of the speech through automatic speech recognition (ASR) \footnote{Automatic speech recognition is limited to the run-time inference phase because in the training phase we directly use the text transcriptions from the training data.} which is convenient for the emotion cues integration in Stage-4. Specifically, we adopt whisper-large-v2 \cite{radford2022robustspeechrecognitionlargescale} as the ASR engine. For Speech Silence Elimination module, we adopt a frame-based detection process implemented with WebRTC \footnote{\url{https://github.com/wiseman/py-webrtcvad}}, in which the speech signal is first divided into short frames, and each frame is classified as speech or non-speech. A sliding-window decision mechanism is then used to aggregate speech-active segments. Meanwhile, we retain the alignment between the processed frames and their original temporal positions, allowing the MTETR module to generate emotion transition timestamps relative to the original speech.

\textbf{Stage-2: Multi-Task Emotion Transition Recognition.}
As shown in the middle part of Figure. \ref{fig:fig3}, the novel Multi-Task Emotion Transition Recognition (MTETR) module aims to capture emotion transition cues, which are all emotional segments and the corresponding timestamp, in whole discourse-level speech. 
Specifically, we first extract emotional representations using Emotion2vec~\cite{ma2023emotion2vecselfsupervisedpretrainingspeech}, and then apply a ResNet~\cite{he2016resnet} to capture local emotional changes. A Transformer~\cite{vaswani2017attention} and a BiLSTM~\cite{graves2012lstm} are used to model long-range contextual emotional dependencies. Two linear layers are employed to jointly perform \textit{Emotion Transition Diarization (ETDia)} and \textit{Emotion Transition Detection (ETDet)}. Note that the ETdia and ETDet are both binary classification tasks at the frame level.
Figure. \ref{fig:fig3} shows the existence of two emotions in the discourse speech, that is, an example of the transition situation in 1.
We can find that the purpose of ETDet is to detect the boundary frames at two different emotional shifting points. We set the two boundary frames to label 1 and the other frames to label 0. The task of ETDia is to predict the duration of different emotions. We set all frames of the first emotion to label 1 and all frames of the second emotion to label 2. 
It is worth noting that the MTETR needs to be pre-trained before the whole pipeline starts. The pre-training data comes from our \textit{EmoTransSpeech}, and the ground truth supervision signals of the multi-task classification loss function are obtained from the duration information of each speech in the data \footnote{Detailed pretraining settings, including dataset, loss functions, etc., are provided in the Appendix A.2.}. 
ETDet is introduced as an auxiliary task during pretraining, which enhances the boundary awareness of MTETR in ETDia. 
After multi-task prediction, we format the module output to the emotion segments or sentences and their annotated emotional timestamps, including the start time, end time, and the corresponding emotion category, 
\color{black}
\underline{e.g. \{start\_time: 00:00, end\_time: 00:05, emotion: "Angry"}; \underline{start\_time: 00:05, end\_time: 00:08, emotion: "Sadness".\} }

\textbf{Stage-3: Acoustic Attributes Analysis.}
Stage-1 and Stage-2 both deal with discourse-level speech, and Stage-3 analyzes the emotional segments isolated from Stage-2, that is, single emotional sentences. We follow SpeechCraft \cite{jin2024speechcraft} to implement the workflow. The output labels consist of speaking style attributes, including pitch, energy, speed, and speaker profile attributes, which include age and gender. For more details, please refer to the SpeechCraft \cite{jin2024speechcraft}. After analyzing each single sentence, we align all the analysis results with each sentence in the discourse-level speech and build an \textit{Emotion Transition-aware Acoustic Attributes} sequence for that discourse.

\textbf{Stage-4: Emotion Cues Integration.} To generate fine-grained, emotion transition-aware descriptions for discourse-level speech, we leverage the strong semantic understanding and text generation capabilities of Gemma-3 \cite{team2025gemma}. Specifically, we integrate the outputs from the preceding three stages: text transcription, emotional timestamps, and emotion transition-aware acoustic attributes. To ensure the applicability and generality of the dataset across various downstream tasks, we construct the \textit{Emotion Cues Integration Prompt} to guide the Gemma-3 to produce two distinct types of captions: \textbf{Descriptive Version (\textit{EmoTransCap ($V_{D}$)})}: This version contains both global and partial descriptions, providing natural language interpretations of discourse-level emotional transitions at both the discourse and segment levels. It is suited for tasks such as speech understanding and multimodal analysis. \textbf{Instructional Version (\textit{EmoTransCap ($V_{I}$)})}: This version is more concise and is formatted using Speech Synthesis Markup Language (SSML) \cite{taylor1997ssml}. It is designed specifically for controllable speech synthesis, enabling models to generate discourse-level speech guided by text-based instructions that reflect emotional transitions:
During \textit{Emotion Cues Integration Prompt} construction, we focus on three key aspects: (1) detecting and avoiding inappropriate or structurally disorganized content, (2) enhancing the diversity and expressiveness of the generated text, and (3) preventing omissions of input labels. To this end, we apply a strict automatic validation mechanism to guide the LLM, filtering out low-quality outputs and triggering regeneration when necessary. This ensures that the final captions meet high standards of accuracy, consistency, and reliability. \textcolor{black}{Please refer to the Appendix A.2 for details on the \textit{Emotion Cues Integration Prompt}}.

\section{Experiments}
\subsection{Experimental Setup}
The experiments are divided into three parts, including (1) Dataset Quality Assessment, (2) Emotion Perception Evaluation, and (3) Emotion Expression Evaluation.

\textbf{(1) Dataset Quality Assessment}: To evaluate the overall quality and practical utility of the proposed EmoTransSpeech ($V_{I}$) dataset, and to ensure the validity of subsequent experiments, we conduct a comprehensive assessment from three perspectives: 1) the quality of the synthesized speech audio, 2) the degree to which the captions accurately reflect emotional transitions, and 3) the semantic consistency between the captions and the corresponding speech content.

\textbf{(2) Emotion Perception Evaluation:} Since all advanced SEC systems, that are {SECap} \cite{xu2023secapspeechemotioncaptioning}, {SpeechCraft} \cite{jin2024speechcraft}, and {AlignCap} etc., share a core scheme, which is to use LLM to read the tokens of the input speech and then output the emotional language description, we therefore choose the open-source {SECap} \cite{xu2023secapspeechemotioncaptioning} system and {SpeechCraft} \cite{jin2024speechcraft} as the most representative baselines.
Specifically, we employ SECap, SpeechCraft, EmoTransCap ($V_{D}$) and EmoTransCap ($V_{I}$) to analyze discourse-level speech and evaluate their emotion perception capabilities from two perspectives: the ability to capture emotional transitions and the semantic consistency between captions and speech. In addition, we fine-tune SECap and SpeechCraft on the EmoTransSpeech dataset to further examine whether the proposed dataset can enhance the perception capabilities of existing SEC models.

\textbf{(3) Emotion Expression Evaluation}: To examine whether fine-tuning on the EmoTransSpeech dataset can enhance the emotional expressiveness of the model, we conduct a comparative experiment between the original {CosyVoice2} without fine-tuning and the version fine-tuned with {EmoTransCap} ($V_{I}$), thereby assessing the contribution of the EmoTransSpeech dataset to improving the ability of TTS models to understand complex emotion prompts.
Additionally, we systematically investigate the impact of two different prompt formats on CosyVoice2's performance:
\textbf{Format-1}: {\texttt{<Speech Caption$_i$>} Text$_i$ \texttt{</Speech>}}, and 
\textbf{Format-2}: {\texttt{Caption <|End-of-prompt|>} Discourse Text}. More experimental setup details can be found in Appendix A.3.

\subsection{Evaluation Metrics}
\textbf{Dataset Quality Assessment. }
\textbf{Subjective Metrics:} 
$\bullet$ \textbf{Emotion Transition Count Accuracy (Acc$_{ETC}$)}. This metric measures whether the number of emotional transitions reflected in the generated captions matches the actual number of transitions in the speech.
$\bullet$ \textbf{Emotion Transition Type Accuracy (Acc$_{ETT}$)}. Given that the predicted number of transitions is correct, this metric further assesses whether the specific types of transitions (e.g., \textit{angry} $\rightarrow$ \textit{sad}) are correctly identified.
$\bullet$ \textbf{Caption-Speech Semantic Consistency Score (MOS-C)}, which evaluates the semantic consistency between the generated captions and the corresponding speech.
$\bullet$ \textbf{Speech Naturalness Score (MOS-S)}, which assesses the naturalness, clarity, and fluency of the speech.
MOS-S and MOS-C are both rated by human evaluators on a scale from 1 to 5.

\textbf{Emotion Perception Evaluation.} We adopt the same set of evaluation metrics introduced in Dataset Quality Assessment, including \textbf{Acc$_{ETC}$}, \textbf{Acc$_{ETT}$}, \textbf{MOS-C}, to assess the ability of different SEC models to generate emotionally aligned and semantically consistent captions from discourse-level speech. In addition, we explored the use of GPT-based automatic evaluation methods as a substitute for human assessment. However, due to high variance and inconsistent results, these methods were ultimately excluded from our analysis. Moreover, unlike some related studies, our current evaluation setup does not involve access to ground-truth captions, which makes it infeasible to apply conventional reference-based automatic metrics such as BLEU or ROUGE.


\begin{table}[t]
  \centering
  \small
  \label{tab:emotion-perception-simplified}
  \resizebox{\columnwidth}{!}{%
      \begin{tabular}{lcccc}
        \toprule
        Lang & Trans Num & Acc$_{\text{ETC}}$\% / Acc$_{\text{ETT}}\%$ $\uparrow$ & MOS-C $\uparrow$ & MOS-S $\uparrow$ \\
        \midrule
        \multirow{3}{*}{Zh} 
          & One   &  100 / 100 & 4.60 & 4.79 \\
          & Two   &  100 / 100 & 4.50 & 4.67 \\
          & Three &  100 / 95.83 & 4.33 & 4.71 \\
        \midrule
        \multirow{3}{*}{En} 
          & One   &  100 / 95.83 & 4.67 & 4.33 \\
          & Two   &  100 / 100 & 4.58 & 4.50 \\
          & Three &  100 / 100 & 4.42 & 4.33 \\
        \bottomrule
      \end{tabular}
   }
  \caption{Evaluation of the EmoTransSpeech dataset quality}
\end{table}


\begin{table*}[ht]
  \centering
  \small
  \vspace{-3mm}
  \begin{tabular}{llcccccc}
    \toprule
    \multirow{2}{*}{Lang} & \multirow{2}{*}{Model}
      & \multicolumn{2}{c}{One Transition} 
      & \multicolumn{2}{c}{Two Transitions} 
      & \multicolumn{2}{c}{Three Transitions} \\
    \cmidrule(lr){3-4} \cmidrule(lr){5-6} \cmidrule(lr){7-8}
    &
      & Acc$_{\text{ETC}}$ / Acc$_{\text{ETT}}$ $\uparrow$ & MOS-C$\uparrow$
      & Acc$_{\text{ETC}}$ / Acc$_{\text{ETT}}$ $\uparrow$ & MOS-C$\uparrow$
      & Acc$_{\text{ETC}}$ / Acc$_{\text{ETT}}$ $\uparrow$ & MOS-C$\uparrow$ \\
    \midrule
    \multirow{5}{*}{Zh} 
      & ${\text{SECap}_{raw}}$         & 0.00 / 0.00 & 1.00 & 0.00 / 0.00 & 1.00 & 0.00 / 0.00 & 1.20 \\
      & ${\text{SECap}_{trained}}$         & 36.84 / 95.00 & 3.90 & 0.00 / 50.00 & 3.10 & 10.00 / 50.00 & 2.40 \\
      & ${\text{SpeechCraft}_{raw}}$   & 0.00 / 0.00 & 3.40 & 0.00 / 0.00 & 3.70 & 0.00 / 0.00 & 4.00 \\
      & ${\text{SpeechCraft}_{trained}}$   & 22.00 / 90.00 & 4.00 & 0.00 / 0.00 & 4.00 & 15.00 / 66.00 & 4.30 \\
      & EmoTransCap ($V_{I}$)  & \textbf{100} / \textbf{100} & \textbf{4.70} & \textbf{100} / \textbf{100} & \textbf{4.40} & \textbf{100} / \textbf{100} & \textbf{4.60} \\
    \midrule
    \multirow{5}{*}{En} 
      & ${\text{SECap}_{raw}}$         & NA / NA & NA & NA / NA & NA & NA / NA & NA \\
      & ${\text{SECap}_{trained}}$         & NA / NA & NA & NA / NA & NA & NA / NA & NA \\
      & ${\text{SpeechCraft}_{raw}}$   & 0.00 / 0.00 & 3.70 & 0.00 / 0.00 & 3.70 & 0.00 / 0.00 & 3.40 \\
      & ${\text{SpeechCraft}_{trained}}$   & 0.00 / 0.00 & 3.70 & 0.00 / 0.00 &3.70 & 0.00 / 0.00 & 3.40 \\
      & EmoTransCap ($V_{I}$)  & \textbf{100} / \textbf{100} & \textbf{4.00} & \textbf{100} / \textbf{100} & \textbf{4.10} & \textbf{100} / \textbf{100} & \textbf{3.90} \\
    \bottomrule
  \end{tabular}
  \caption{Emotion perception performance of different models across languages and varying numbers of emotional transitions.}
  \label{tab:emotion-perception}
\end{table*}

\begin{table*}
\centering
\small
\begin{tabular}{llccccc}
\toprule
Lang & Method 
& EES$_{ET}^1$~$\uparrow$ & EES$_{ET}^2$~$\uparrow$ & EES$_{ET}^3$~$\uparrow$ & MOS-E~$\uparrow$ & MOS-S~$\uparrow$
\\
\midrule

\multirow{3}{*}{Zh}
& CosyVoice2 w/o Fine-tuning                    & 52.09 & 31.03 & 19.74 & 2.25 & 1.75 \\
& CosyVoice2 w/ EmoTransCap ($V_{I}$) (Format1) & \textbf{68.62} & \textbf{54.61} & \textbf{42.79} & \textbf{4.72} & \textbf{4.33} \\
& CosyVoice2 w/ EmoTransCap ($V_{I}$) (Format2) & 65.61 & 53.55 & 42.21 & 4.46 & 4.00 \\

\midrule

\multirow{3}{*}{En}
& CosyVoice2 w/o Fine-tuning                    & 70.64 & 48.12 & 31.99 & 2.36 & 1.83 \\
& CosyVoice2 w/ EmoTransCap ($V_{I}$) (Format1) & \textbf{73.24} & \textbf{61.47} & \textbf{49.16} &\textbf{ 4.88} & \textbf{4.58}\\
& CosyVoice2 w/ EmoTransCap ($V_{I}$) (Format2) & 70.91 & 60.04 & 46.93 & 4.62 & 4.16\\

\bottomrule
\end{tabular}

\caption{Comparative evaluation of emotional expressiveness in TTS under different fine-tuning and prompt conditions.}
\vspace{-2mm}
\label{tab:tts_score}
\end{table*}


\textbf{Emotion Expression Evaluation.} 
\textbf{1) Subjective Metrics:} 
$\bullet$ \textbf{Emotional Consistency Score (MOS-E)}. This metric evaluates whether the emotional semantics conveyed by the synthesized speech align with the corresponding generated caption. 
$\bullet$ \textbf{Speech Naturalness Score (MOS-S)}. This metric assesses the naturalness, clarity, and fluency of the synthesized speech.
\textbf{2) Objective Metrics:} 
$\bullet$ \textbf{Emotion Embedding Similarity (EES)}. EES$_{ET}^1$, EES$_{ET}^2$ and EES$_{ET}^3$, which respectively measure the similarity between the emotion embeddings of the synthesized speech and the ground-truth speech when expressing one, two, and three emotional transitions.

\section{Results and Discussions}

\subsection{Dataset Quality Assessment}
We randomly selected 30 samples from the EmoTransSpeech ($V_{I}$) dataset for quality evaluation, evenly covering speech segments with one, two, and three emotional transitions. In the subjective evaluation, 20 volunteers were invited to manually assess the samples based on a standardized evaluation guideline. All evaluators were undergraduate or graduate students engaged in speech synthesis research and possessed a high level of domain expertise. The evaluation protocol is detailed in the Appendix A.4. All results are summarized in Table 1. 

The results demonstrate that the EmoTransSpeech ($V_{I}$) dataset performs strongly across all evaluation metrics. It exhibits high-quality synthesized speech, generates captions that accurately reflect the emotional dynamics of the audio, and maintains a high degree of semantic consistency between the captions and the speech content.

\subsection{Emotion Perception Evaluation}
We also randomly selected 30 speech samples from the EmoTransSpeech ($V_{I}$)dataset and invited 20 volunteers to participate in a subjective evaluation. The compared models include the original, non-fine-tuned versions of SECap and SpeechCraft (denoted as ${\text{SECap}_{raw}}$ and ${\text{SpeechCraft}_{raw}}$, respectively), as well as their counterparts fine-tuned on EmoTransSpeech (denoted as ${\text{SECap}_{ft}}$ and ${\text{SpeechCraft}_{ft}}$). We conducted a comprehensive comparison against EmoTransCap ($V_{I}$). The final evaluation results are presented in Table 2.

The results show that EmoTransCap ($V_{I}$) significantly outperforms all other models across the three evaluation dimensions, further demonstrating its superior capability in perceiving dynamic emotional transitions in discourse-level speech and generating more expressive and emotionally aligned caption texts. In addition, we observed that both ${\text{SECap}_{ft}}$ and ${\text{SpeechCraft}_{ft}}$ consistently outperformed their original counterparts across all evaluation metrics. This highlights the significant contribution of the EmoTransSpeech dataset in enhancing the emotion perception capabilities of existing SEC models.

\subsection{Emotion Expression Evaluation}
In this section, we select 1,500 samples for each language (Chinese and English) from the EmoTransSpeech test set and conduct controllable speech synthesis using both the original CosyVoice2 model without fine-tuning (CosyVoice2 w/o Fine-tuning) and the version fine-tuned on the EmoTransSpeech dataset (CosyVoice2 w/ EmoTransCap ($V_{I}$)). Each synthesis is conditioned on both the caption and the corresponding text. The results are summarized in Table 3.

As shown in the results, the original CosyVoice2 model consistently underperforms its fine-tuned counterpart across all metrics, primarily due to its limited ability to interpret complex emotional prompts. In contrast, training on the EmoTransSpeech dataset significantly enhances the model’s capability to understand and model complex emotional instructions, further validating the practical utility of the dataset for controllable emotional speech synthesis. In addition, the results demonstrate that Format-1 consistently outperforms Format-2, regardless of whether the input captions are generated by SECap or EmoTransSpeech ($V_{I}$). A plausible explanation is that Format-1 enables more fine-grained emotional control by establishing a clearer one-to-one correspondence between each caption and its associated text segment. This structural alignment facilitates more precise emotional conditioning during synthesis. In contrast, Format-2 treats the caption as a global prompt for the entire discourse, which may result in weaker alignment with the specific emotional context of individual utterances, thereby reducing the effectiveness of emotion guidance.

\section{Conclusion}
In this paper, we introduce \textbf{EmoTransCap}, a novel paradigm designed for capturing and describing dynamic emotional transitions in discourse-level speech. We present \textbf{EmoTransSpeech}, the first large-scale bilingual dataset specifically developed to address this task, along with a fully automated pipeline capable of generating emotionally aligned captions and expressive speech synthesis. Comprehensive experiments demonstrate that EmoTransCap significantly enhances both emotion perception and emotional expressiveness in SEC and TTS tasks.
However, our work currently relies heavily on synthetic data, which may not fully replicate the complexity of naturally-occurring emotional discourse. Moreover, our automated evaluation remains limited, emphasizing a need for more robust and reliable assessment methods.
Future research will focus on incorporating more naturalistic discourse data, developing improved automatic evaluation techniques, and exploring multimodal cues (e.g., facial and physiological signals) to achieve more comprehensive and human-like emotional understanding and generation.

\section{Acknowledgments}
This research of Rui Liu was funded by the General Program (No.62476146) of the National Natural Science Foundation of China, the Young Elite Scientists Sponsorship Program by CAST (2024QNRC001), the Outstanding Youth Project of Inner Mongolia Natural Science Foundation (2025JQ011), the Key R\&D and Achievement Transformation Program of Inner Mongolia Autonomous Region (2025YFHH0014), the Central Government Fund for Promoting Local Scientific and Technological Development (2025ZY0143).

\bibliography{aaai2026}

\appendix

\section{Technical Appendix}

\section{A.1 EmoTransSpeech Dataset Details}
\begin{figure}[h]
    \centering
    \begin{subfigure}{0.4\textwidth}
        \centering
        \includegraphics[width=1\linewidth]{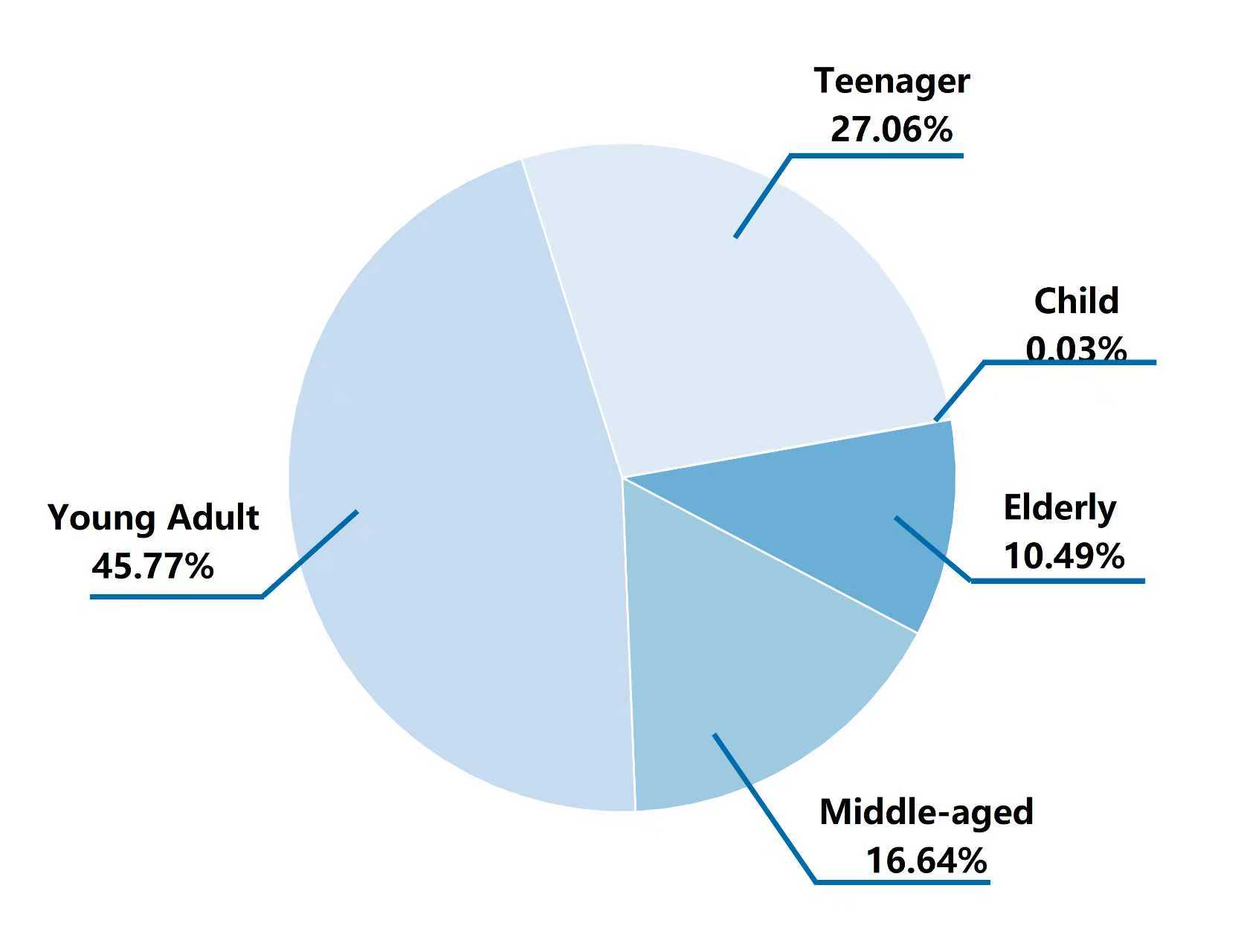}
        \label{fig:age}
    \end{subfigure}%
    \hfill
    \caption{Distribution of age}
    \label{fig:Age}
\end{figure}

Table \ref{tab:dataset_stats} presents recent speech-caption paired datasets. As shown, EmoTransSpeech is the only bilingual dataset (English and Chinese) that supports both discourse-level modeling and emotion transition annotation. Although it is smaller in scale compared to SpeechCraft, it offers more targeted and task-adaptive emotional structure modeling and fine-grained caption customization. Table \ref{tab:statistic} presents the detailed statistics of EmoTransSpeech. Additionally, we have also analyzed the age distribution in EmoTransSpeech, with the results shown in Figure \ref{fig:Age}. 

\begin{figure}[h]
    \centering
    \includegraphics[width=1\linewidth]{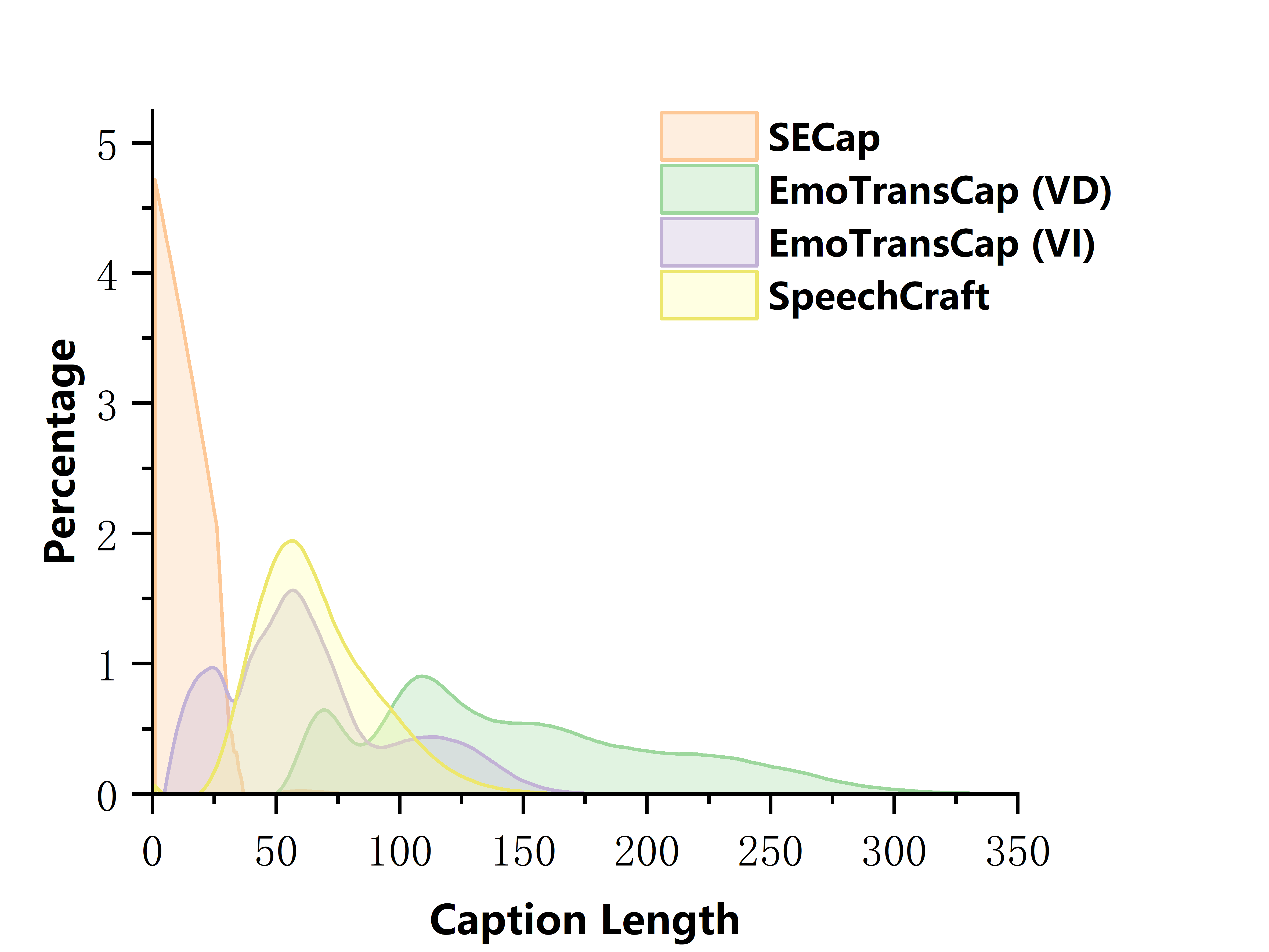}
    \caption{Distribution of Caption Lengths Across Different Models}
    \label{fig:caption_length}
\end{figure}

Figure \ref{fig:caption_length} illustrates the distribution of caption lengths generated by different models. As shown in the plot, SECap generates relatively short captions, with most of them containing fewer than 25 words, reflecting the model's tendency to produce concise descriptions. On the other hand, SpeechCraft, which incorporates additional fine-grained information such as age, gender, and tone, generates longer captions with more detailed content. Despite the increased information, the length distribution of SpeechCraft remains relatively concentrated. In contrast, EmoTransCap (VI) and EmoTransCap (VD), which integrate the Multi-Task Emotion Transition Recognition (MTETR) module, provide finer-grained captions for speech containing multiple emotional transitions. This leads to a more uniform distribution of caption lengths, indicating the models' ability to generate detailed descriptions across a wider range of lengths, capturing the complexity of emotional shifts in discourse.

\setlength{\tabcolsep}{8pt}
\renewcommand{\arraystretch}{1.2}

\begin{figure*}[t]
  \centering
  \begin{subfigure}[b]{0.24\textwidth}
    \includegraphics[width=\linewidth]{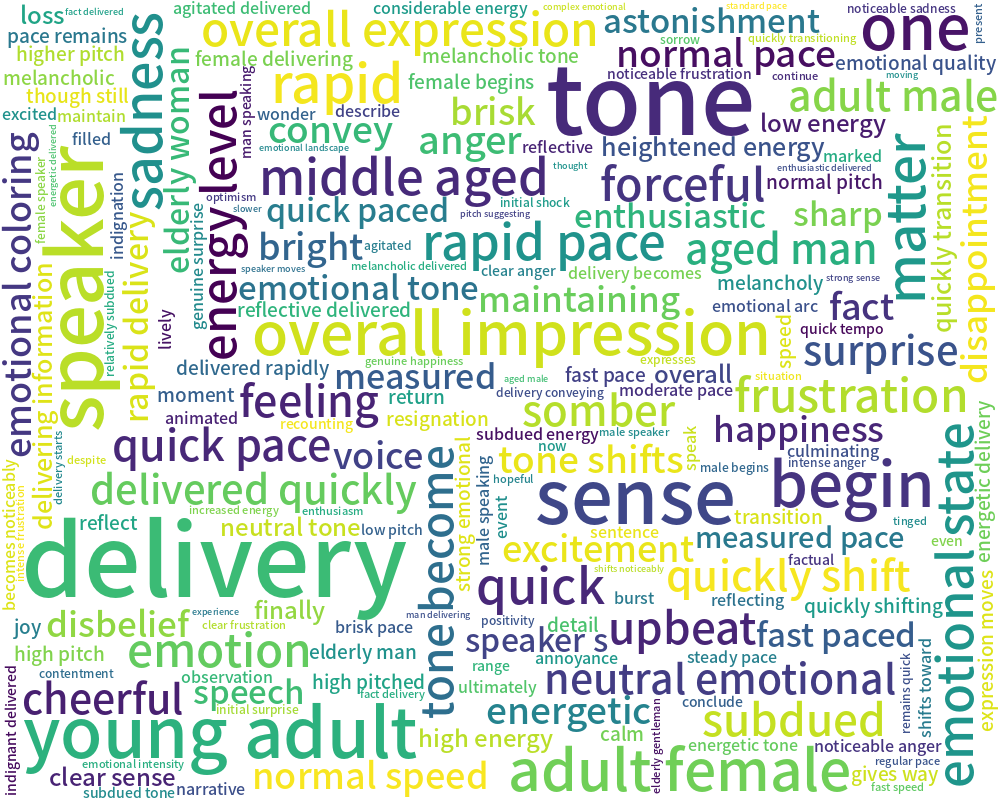}
    \caption{English ($V_{D}$)}
    \label{fig:desc_en}
  \end{subfigure}\hfill
  \begin{subfigure}[b]{0.24\textwidth}
    \includegraphics[width=\linewidth]{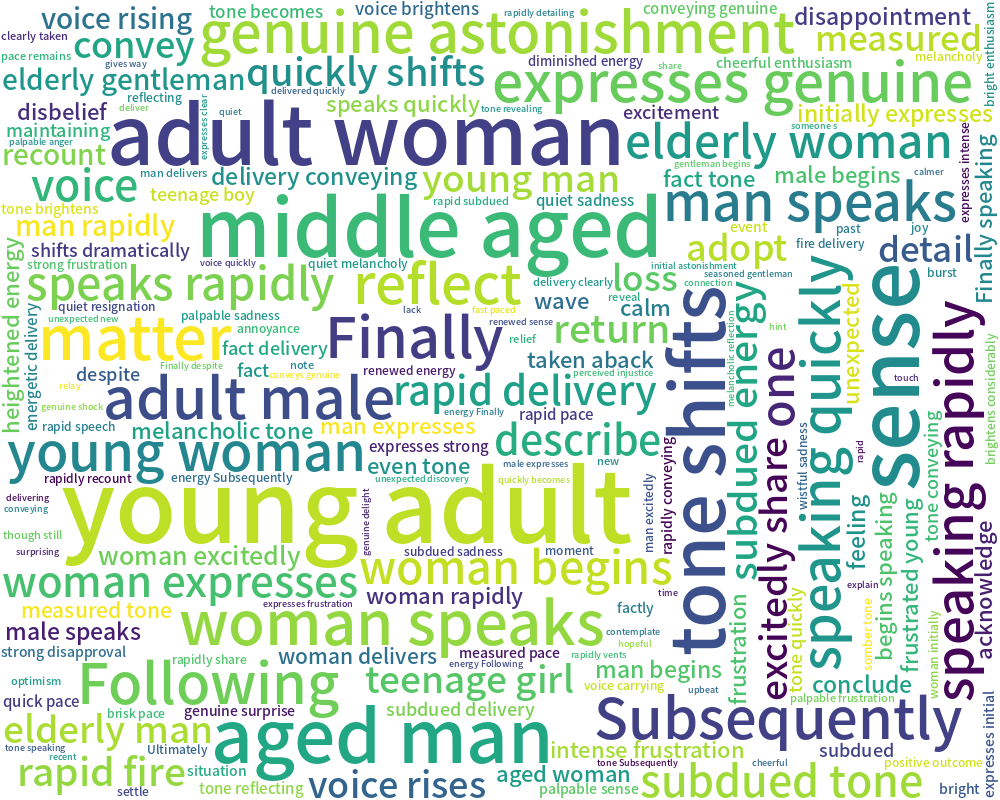}
    \caption{English ($V_{I}$)}
    \label{fig:instr_en}
  \end{subfigure}\hfill
  \begin{subfigure}[b]{0.24\textwidth}
    \includegraphics[width=\linewidth]{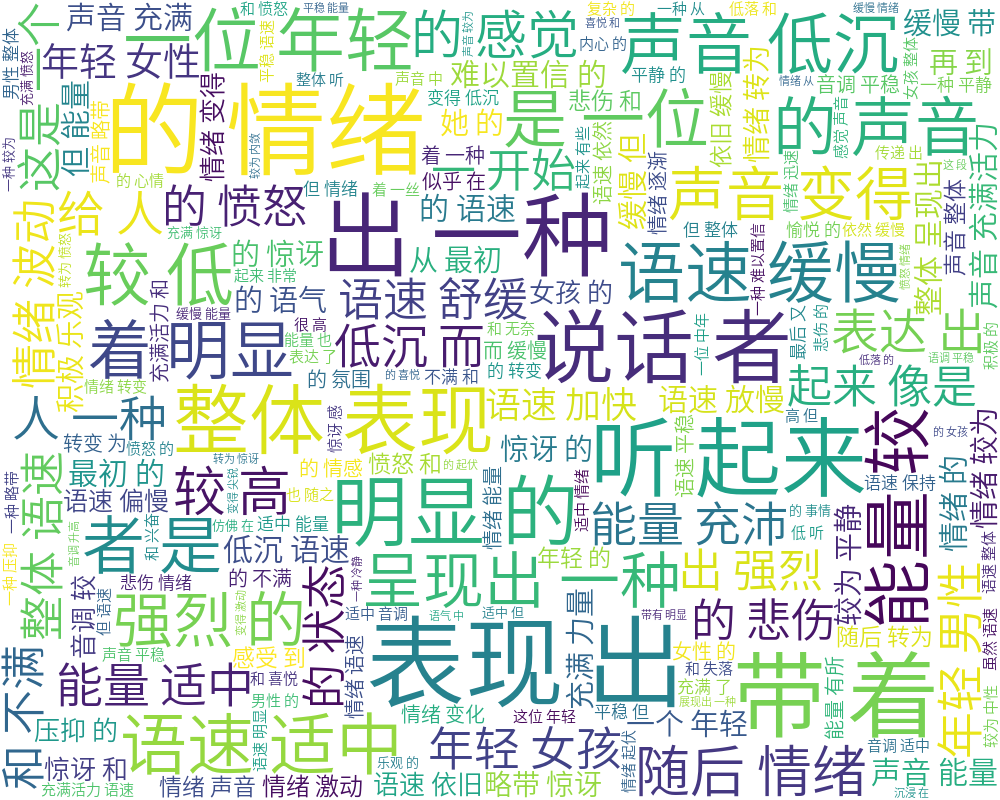}
    \caption{Chinese ($V_{D}$)}
    \label{fig:desc_zh}
  \end{subfigure}\hfill
  \begin{subfigure}[b]{0.24\textwidth}
    \includegraphics[width=\linewidth]{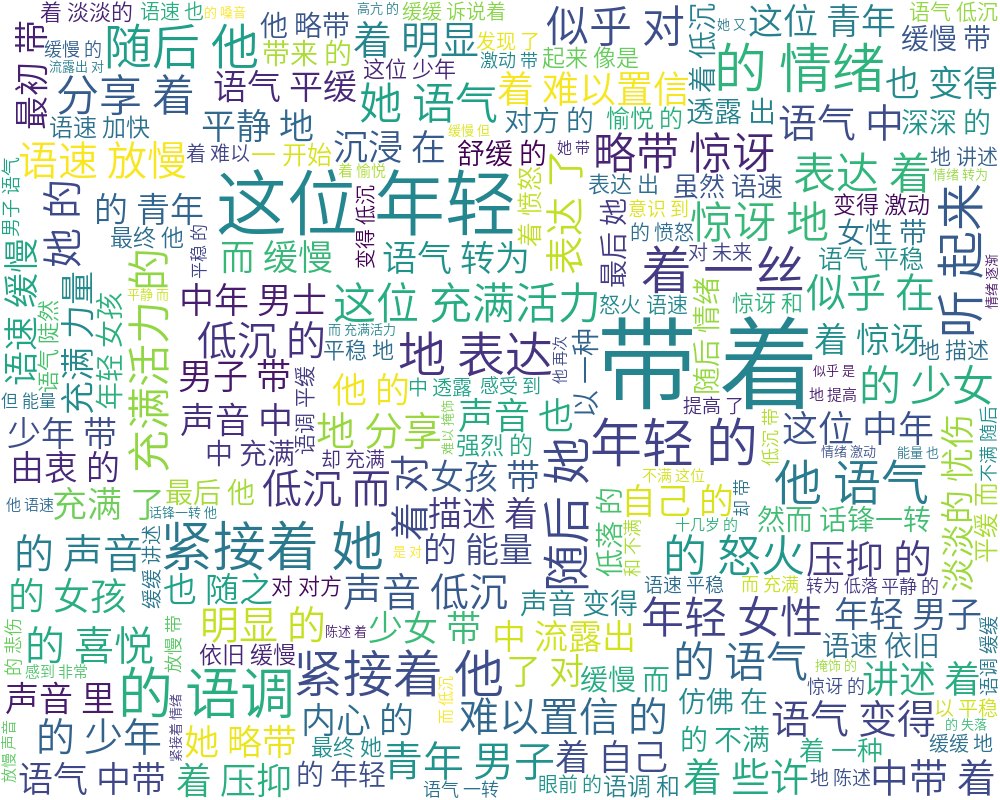}
    \caption{Chinese ($V_{I}$)}
    \label{fig:instr_zh}
  \end{subfigure}
  \caption{Word clouds of captions.}
  \label{fig:wordcloud_all}
\end{figure*}

\begin{table*}[t]
\centering
\small
\resizebox{\textwidth}{!}{ 
\begin{tabular}{@{}lcccccccc@{}} 
\toprule
\textbf{Dataset} & \textbf{Duration} & \textbf{Clips} & \textbf{Speakers} & \textbf{Caption Form}  & \textbf{Language} & \textbf{Discourse-Level} & \textbf{Emotion Transition} \\
\midrule
PromptSpeech   & /     &   28,000   &    /    & Style tag &    EN       & $\times$ & $\times$ \\
TextrolSpeech  & 330h  &   236,000   &  1,324      &  LLM template   &    EN       & $\times$ & $\times$ \\ 
ZED            & 17min &   180   &   73     &   /  &     EN       & $\times$ & $\times$ \\
SpeechCraft    & 2,391h &  2,250,000   &   $>$3,200     &   LLM customization   &    EN + ZH  & $\times$ & $\times$ \\
\midrule
\textbf{EmoTransSpeech (ours)}  & 617h  &  144,000    &    20    &   LLM customization   &                        EN + ZH  & $\checkmark$ & $\checkmark$ \\
\bottomrule
\end{tabular}
} 
\caption{Dataset Statistics}
\label{tab:dataset_stats}
\end{table*}

\begin{table*}[t]
\centering
\small
\begin{tabular}{@{}c!{\vrule width 0.4pt}
                cc!{\vrule width 0.4pt}
                cc!{\vrule width 0.4pt}
                cc!{\vrule width 0.4pt}
                cc@{}}
\toprule
\textbf{Item}                         & \multicolumn{2}{c!{\vrule width 0.4pt}}{\textbf{w/o Trans}} & \multicolumn{2}{c!{\vrule width 0.4pt}}{\textbf{One Trans}} & \multicolumn{2}{c!{\vrule width 0.4pt}}{\textbf{Two Trans}} & \multicolumn{2}{c}{\textbf{Three Trans}} \\ 
\midrule
Language                              & $EN$     & $ZH$     & $EN$     & $ZH$     & $EN$     & $ZH$     & $EN$     & $ZH$     \\
Utterances                            & 20,000   & 20,000   & 20,000   & 20,000   & 16,000   & 16,000   & 16,000   & 16,000   \\
Words                                 & 408,791  & 487,603  & 661,461  & 971,010  & 631,980  & 959,681  & 749,831  & 1,171,623\\
Max words per utterance               & 40       & 40       & 76       & 80       & 78       & 120      & 89       & 159      \\
Min words per utterance               & 5        & 5        & 11       & 12       & 17       & 17       & 22       & 21       \\
Mean words per utterance              & 20       & 24       & 33       & 48       & 39       & 60       & 47       & 73       \\
\midrule
Duration(h)                           & 51.21    & 35.51    & 90.41    & 72.81    & 91.05    & 73.53    & 111.89   & 91.21    \\
Max utterance duration(s)             & 20.92    & 18.16    & 34.56    & 27.88    & 38.08    & 40.88    & 44.96    & 51.84    \\
Min utterance duration(s)             & 2.04     & 1.40     & 5.36     & 3.72     & 8.72     & 5.24     & 12.20    & 7.56     \\
Mean utterance duration(s)            & 9.22     & 6.39     & 16.27    & 13.11    & 20.49    & 16.54    & 25.17    & 20.52    \\
\midrule
Max caption ($V_{I}$) length          & 34       & 96       & 55       & 114      & 80       & 169      & 95       & 206      \\
Min caption ($V_{I}$) length          & 12       & 32       & 19       & 45       & 29       & 65       & 41       & 78       \\
Mean caption ($V_{I}$) length         & 19       & 52       & 35       & 73       & 51       & 104      & 64       & 128       \\
\midrule
Max caption ($V_{D}$) length          & 109      & 193      & 153      & 283      & 189      & 337      & 215      & 392      \\
Min caption ($V_{D}$) length          & 41       & 57       & 64       & 94       & 83       & 129      & 98       & 156      \\
Mean caption ($V_{D}$) length         & 65       & 108      & 99       & 174      & 121      & 213      & 146      & 249      \\
\midrule
Emotion Transitions                   & 5        & 5        & 20       & 20       & 80       & 80       & 320      & 320      \\
Speakers                              & 10       & 10       & 10       & 10       & 10       & 10       & 10       & 10       \\
\bottomrule
\end{tabular}
\caption{\label{tab:statistic}\textcolor{black}{The detailed data statistics of EmoTransSpeech.}} 
\end{table*}

\section{A.2 EmoTransCap Pipeline Details}
\subsection{A.2.1 Pretraining MTETR Module}
During training, Emotion Transition Diarization (ETDia) was defined as the primary task, while Emotion Transition Detection (ETDet) served as an auxiliary task. The overall model architecture comprises a ResNet, a Transformer, and several linear layers. Detailed parameter settings are presented in Table \ref{tab:MTETR_Para}. Notably, we introduced an Uncertainty Loss to dynamically balance the learning process of these two tasks, defined as follows:
\begin{equation}
    \small
    \mathcal{L}_{\text{total}} = \frac{1}{2\sigma_{\text{ETDia}}^2} \mathcal{L}_{\text{ETDia}} + \log \sigma_{\text{ETDia}} + \frac{1}{2\sigma_{\text{ETDet}}^2} \mathcal{L}_{\text{ETDet}} + \log \sigma_{\text{ETDet}}
\end{equation}





To comprehensively assess the effectiveness of the auxiliary task and individual structural components, we conducted a series of ablation studies. Specifically, under the multi-task setting \footnote{Both tasks were trained simultaneously}, we constructed four model variants by selectively removing zero, one, or both of the ResNet and Transformer modules, and examined their impact on performance. Furthermore, to evaluate the contribution of the auxiliary task, we retrained the same four model variants under the single-task setting \footnote{Only the ETDia objective was optimized}, in which the ETDet objective was removed. Importantly, to ensure fair comparisons, all model variants were constrained to have approximately the same number of parameters as the full model.

In addition, to evaluate the performance of the MTETR model on realistic discourse-level emotional transition speech, we constructed a dedicated test set totaling approximately 6 hours in duration, based on the ESD dataset. It is worth noting that all speech samples included in this test set were strictly excluded from being used as reference samples during \textit{Stage-2: Speech Generation}, thereby ensuring a fair and unbiased evaluation protocol. As shown in Table \ref{tab:MTETR_Result}, removing either the ResNet or Transformer component leads to a noticeable performance drop, indicating that both structural modules contribute significantly to the model’s effectiveness. Furthermore, eliminating the auxiliary task under the single-task setting also results in degraded performance. These findings demonstrate the effectiveness of our proposed multi-task architecture, which enhances the model’s ability to capture complex emotional dynamics in discourse-level speech.

\subsection{A.2.2 Emotion Cues Integration Prompt}
To generate detailed and accurate emotion transition-aware captions, we developed two specialized prompts utilized in \textit{Stage-4: Emotion Cues Integration}. These prompts guide Gemma-3 to produce distinct caption styles—Descriptive Version (EmoTransCap ($V_{D}$)) and Instructional Version (EmoTransCap ($V_{I}$)), as described in the main paper.

\textbf{(1) Prompt for EmoTransCap ($V_{I}$)}. This prompt is designed to guide the model in generating concise, expressive, and objective captions to support controllable speech synthesis tasks.
\begin{quote}
\textbf{Below is an instruction that describes a task. Write a response that appropriately completes the request.}

\textbf{Instruction:}  \\
You will be provided with paralinguistic attributes of an audio clip containing \{segment\_num\} consecutive segment(s) spoken by the same speaker. Your task is to generate \{segment\_num\} fluent, expressive, and objective caption(s), each describing the speaker's vocal style, emotional dynamics, and communicative intent.

\textbf{Output Guidelines:}
\begin{enumerate}
\item Output exactly \{segment\_num\} line(s)—one caption per segment—without numbering.
\item Do NOT quote or reuse any transcript content.
\item The language must reflect attributes such as speaking rate, pitch, energy, and emotion in a natural and concise manner.
\item Mention the speaker's gender and age only in the first caption, embedding them naturally into the sentence.
\item Ensure coherence between captions using appropriate transition words.
\item The tone must remain descriptive and objective.
\end{enumerate}

\textbf{Input:}  \\
\{segment\_descriptions\}

\end{quote}

\textbf{(2) Prompt for EmoTransCap ($V_{D}$)}. This prompt instructs the model to generate both global and segment-level descriptions, providing detailed and natural language interpretations suitable for speech understanding and multimodal analysis tasks:

\begin{quote}
\textbf{Below is an instruction that describes a task. Write a response that appropriately completes the request.}

\textbf{Instruction:}  \\
You will be given an audio clip divided into \{segment\_num\} segment(s), each containing pitch, speed, energy, emotion, age, gender, transcript, and start/end timestamps. Based on this information, generate two parts:

\textbf{[Global Description]}
Provide an overall natural-language description of the speaker’s emotional state, vocal tone, and prosodic variation across the entire audio clip.

\textbf{[Partial Description]}
Provide a short and objective description for each segment individually.

\textbf{Output Guidelines:}

\textbf{[Global Description]}
\begin{enumerate}
\item Describe emotional dynamics, tone shifts, speaking rate, and pitch variations throughout the full clip.
\item Integrate gender and age naturally.
\item Reference the transcript for context, but do not copy or quote any part of it.
\item Use fluent, concise, and descriptive language. Avoid excessive sentiment, symbolic characters (e.g., *, \#), or line breaks.
\item If there is only one segment, do not mention emotional changes.
\end{enumerate}

\textbf{[Partial Description]}
\begin{enumerate}
\item Each segment must begin with the format: \textbf{PartX (start\_time \textasciitilde{} end\_time)}.
\item Use full sentences to objectively describe pitch, speed, energy, and emotion in the segment.
\item Do not refer to the speaker or use subjective terms; keep each description self-contained.
\item Reference the transcript but do not quote it.
\item Descriptions should be short, fluent, and symbol-free.
\end{enumerate}

\textbf{Input:}  \\
\{segment\_data\}

\end{quote}

These two prompts were carefully designed to ensure high standards of caption accuracy, consistency, and expressiveness, thereby maximizing the utility and applicability of the generated dataset for diverse downstream speech analysis and synthesis tasks.

\section{A.3 Experiment Setup Details}

\subsection{A.3.1 Fine-tuning CosyVoice2}

In addition to the experiments described in the main text, we also utilize two other baseline models, namely SECap (w/o fine-tuning) and SpeechCraft (w/o fine-tuning), to independently generate captions for the EmoTransSpeech-Audio dataset. The generated audio, captions, and corresponding texts are then used for fine-tuning the CosyVoice2 model. Furthermore, considering that the original CosyVoice2 can inherently learn emotional transitions from discourse-level text, we also remove the captions \textbf{(w/o caption)} to evaluate the benefits of caption-guided supervision. Additionally, we evaluate the models using three objective metrics (\textbf{EES$_{ET}^{1\rightarrow 3}$}), and the results are shown in Table \ref{tab:CosyVoice2_Result}.

For specific parameter settings of the CosyVoice2 model, please refer to Table \ref{tab:CosyVoice2_Para}. Additional details regarding the model configuration can be found in the code submitted along with this paper. Moreover, the Fine-tuning of CosyVoice2 was conducted on a single NVIDIA A800 GPU.

From the EmoTransSpeech-Audio dataset, we select a subset of 104 hours of audio data, splitting it into training and validation sets with a ratio of 3.5:1. The test set comprises an independent collection of 3,000 samples (1,500 samples each for Chinese and English).



\subsection{A.3.2 Fine-tuning SECap}  
We selected 59,515 audio-caption pairs from the EmoTransSpeech dataset to fine-tune the SECap model. The fine-tuning experiments strictly followed the official implementation, using a batch size of 16 across three NVIDIA A800 GPUs. To ensure a fair evaluation of emotion perception performance, we explicitly excluded all speech samples present in the test set from the SECap training data.

\renewcommand{\arraystretch}{1.2} 
\begin{table}[t]
  \centering
  \small
  \label{tab:mtetr}
  \begin{tabular}{@{}lll@{}}
    \toprule
    \textbf{Module} & \textbf{Parameter}         & \textbf{Value} \\ 
    \midrule
    \multirow{6}{*}{ResNet} 
      & res\_blocks                & 8     \\ 
      & in\_planes                 & 768   \\ 
      & planes                     & 512   \\ 
      & embed\_dim                 & 128   \\ 
      & kernel\_size               & 1     \\ 
      & stride                     & 1     \\ 
    \midrule
    \multirow{5}{*}{Transformer} 
      & num\_layers                & 2     \\ 
      & nhead                      & 4     \\ 
      & d\_model                   & 128   \\ 
      & dim\_feedforward           & 1,024  \\ 
      & dropout                    & 0.5   \\ 
    \midrule
    \multirow{4}{*}{Linear} 
      & in\_features               & 128   \\ 
      & hidden\_features           & 256 \\ 
      & out\_features (ETDet)      & 1     \\ 
      & out\_features (ETDia)      & 5     \\ 
    \bottomrule
  \end{tabular}
  \caption{MTETR Model Architecture Parameters}
  \label{tab:MTETR_Para}
\end{table}

\begin{table}[t]
  \centering
  \small
  \label{tab:cosyvoice2}
  \begin{tabular}{@{}lll@{}}
    \toprule
    \textbf{Category}   & \textbf{Parameter}              & \textbf{Value} \\ 
    \midrule
    \multirow{4}{*}{Normal} 
      & max\_frames\_in\_batch    & 1,500  \\ 
      & speech\_sample\_rate       & 24,000 \\ 
      & seed                       & 1,986  \\ 
      & lr                         & 1e-5  \\ 
    \midrule
    \multirow{3}{*}{LLM} 
      & llm\_input\_size           & 896   \\ 
      & llm\_output\_size          & 896   \\ 
      & speech\_token\_size        & 6,561  \\ 
    \midrule
    \multirow{4}{*}{Sampling} 
      & top\_p                     & 0.8   \\ 
      & top\_k                     & 25    \\ 
      & tau\_r                     & 0.1   \\ 
      & win\_size                  & 10    \\ 
    \midrule
    \multirow{4}{*}{Flow} 
      & flow\_input\_size          & 512   \\ 
      & flow\_output\_size         & 80    \\ 
      & spk\_embed\_dim            & 0.1   \\ 
      & vocab\_size                & 6,561  \\ 
    \bottomrule
  \end{tabular}
  \caption{CosyVoice2 Hyperparameter Configuration}
  \label{tab:CosyVoice2_Para}
\end{table}

\begin{table*}
\centering
\small
\begin{tabular}{llccc}
\toprule
\textbf{Lang} & \textbf{Method} 
& \textbf{ETT$_{ET}^1$}~$\uparrow$ & \textbf{ETT$_{ET}^2$}~$\uparrow$ & \textbf{ETT$_{ET}^3$}~$\uparrow$ \\
\midrule

\multirow{6}{*}{Zh}
& CosyVoice2 w/o Caption             & 60.03 & 46.31 & 35.87  \\
& CosyVoice2 w/ SpeechCraft           & 58.54 & 42.86 & 31.01 \\
& CosyVoice2 w/ SECap (Format1)      & 61.77 & 48.35 & 35.83  \\
& CosyVoice2 w/ SECap (Format2)      & 61.77 & 46.50 & 34.85  \\
& CosyVoice2 w/ EmoTransCap ($V_{I}$) (Format1) & \textbf{68.62} & \textbf{54.61} & \textbf{42.79} \\
& CosyVoice2 w/ EmoTransCap ($V_{I}$) (Format2) & 65.61 & 53.55 & 42.21 \\

\midrule

\multirow{6}{*}{En}
& CosyVoice2 w/o Caption             & 69.87 & 57.70 & 45.73 \\
& CosyVoice2 w/ SpeechCraft           & 68.99 & 57.16 & 44.89 \\
& CosyVoice2 w/ SECap (Format1)      & NA    & NA    & NA   \\
& CosyVoice2 w/ SECap (Format2)      & NA    & NA    & NA    \\
& CosyVoice2 w/ EmoTransCap ($V_{I}$) (Format1) & \textbf{73.24} & \textbf{61.47} & \textbf{49.16}\\
& CosyVoice2 w/ EmoTransCap ($V_{I}$) (Format2) & 70.91 & 60.04 & 46.93 \\

\bottomrule
\end{tabular}

\caption{Emotion Expression of EmoTransCap}
\label{tab:CosyVoice2_Result}
\end{table*}

\begin{table*}[t]
  \label{tab:MTETR}
  \centering
  \small
  \begin{tabular}{c!{\vrule width 0.4pt}c!{\vrule width 0.4pt}ccc!{\vrule width 0.4pt}ccccc}
    \toprule
    \textbf{Lang} & \textbf{Task} & \textbf{ResNet} & \textbf{Transformer} & \textbf{Linear} 
    & \textbf{Acc$_{ET}^1$~$\uparrow$} & \textbf{Acc$_{ET}^2$~$\uparrow$} & \textbf{Acc$_{ET}^3$~$\uparrow$} & \textbf{FEA~$\uparrow$} & \textbf{EER~$\downarrow$} \\
    \midrule

    \multirow{8}{*}{Zh} & \multirow{4}{*}{Single Task}
     & \CheckmarkBold & \CheckmarkBold & \CheckmarkBold & \textbf{49.28} & \textbf{39.43} & \textbf{26.43} & \textbf{74.03} & NA \\
    & &   & \CheckmarkBold  & \CheckmarkBold & 38.28 & 29.71 & 24.71 & 69.39 & NA \\
    & & \CheckmarkBold &   & \CheckmarkBold & 22.43 & 22.00 & 26.43 & 68.65 & NA \\
    & &   &   & \CheckmarkBold & 18.14 & 15.71 & 14.71 & 61.28  & NA \\
    \cmidrule(lr){2-10}

    &\multirow{4}{*}{Multi Task}
     & \CheckmarkBold & \CheckmarkBold & \CheckmarkBold & \textbf{50.00} & \textbf{41.57} & \textbf{30.57} & \textbf{75.37} & \textbf{5.43} \\
    & &   & \CheckmarkBold & \CheckmarkBold & 40.14 & 34.43 & 28.71 & 71.53 & 6.81 \\
    & & \CheckmarkBold &   & \CheckmarkBold & 22.71 & 25.14 & 27.43 & 69.22 & 6.33 \\
    & &   &   & \CheckmarkBold & 19.43 & 16.14 & 16.14 & 61.88 & 8.07 \\
    \midrule
    
    \multirow{8}{*}{En} & \multirow{4}{*}{Single Task}
     & \CheckmarkBold & \CheckmarkBold & \CheckmarkBold & \textbf{19.38} & \textbf{9.75} & \textbf{5.12} & \textbf{48.23} & NA \\
    & &   & \CheckmarkBold & \CheckmarkBold & 10.50 & 4.62 & 2.12 & 42.41 & NA \\
    & & \CheckmarkBold &   & \CheckmarkBold & 7.88 & 5.88 & 5.00 & 43.78 & NA \\
    & &   &  & \CheckmarkBold & 7.75 & 4.25 & 3.12 & 43.50 & NA \\
    \cmidrule(lr){2-10}
    
     &\multirow{4}{*}{Multi Task}
     & \CheckmarkBold & \CheckmarkBold & \CheckmarkBold & \textbf{21.12} & \textbf{10.12} & \textbf{4.75} & \textbf{49.88} & \textbf{9.43} \\
     & & & \CheckmarkBold & \CheckmarkBold & 9.75 & 3.50 & 3.12 & 42.81 & 13.25 \\
     & & \CheckmarkBold & & \CheckmarkBold & 8.62 & 5.38 & 4.75 & 44.00 & 10.50 \\
     & &   &   & \CheckmarkBold & 7.76 & 4.88 & 2.00 & 41.72 & 13.26 \\
    \bottomrule
    
  \end{tabular}%
  \caption{The fine-grained ablation experiment results for the Multi-Task Emotion Transition Recognition (MTETR)}
  \label{tab:MTETR_Result}
\end{table*}

\section{A.4 Evaluation Details}

\subsection{A.4.1 User Evaluation Instructions}
\textbf{MOS-S}. Please assess the audio quality of each synthesized speech sample using the following 5-point rating scale:
\begin{itemize}
    \item \textbf{5 points:} Speech is natural, fluent, clear, and stable; nearly indistinguishable from human speech.
    \item \textbf{4 points:} Speech is mostly natural with minor and non-intrusive synthesis artifacts that do not impact comprehension or listening experience.
    \item \textbf{3 points:} Noticeable synthesis artifacts, including unnatural pauses or fluctuations in pitch/timbre; however, speech remains semantically clear and comprehensible.
    \item \textbf{2 points:} Speech is noticeably unnatural with significant audio quality issues such as noise, interruptions, or prolonged pronunciations, causing moderate comprehension difficulty.
    \item \textbf{1 point:} Speech is unintelligible, with extremely poor audio quality, making the content essentially incomprehensible.
\end{itemize}

\textbf{MOS-C}. Evaluate the semantic consistency between each caption and its corresponding synthesized speech sample using the scale below:
\begin{itemize}
    \item \textbf{5 points:} Completely consistent (fully relevant).
    \item \textbf{3 points:} Partially consistent (some relevance).
    \item \textbf{1 point:} Completely inconsistent (no relevance).
\end{itemize}

\textbf{MOS-E}. Evaluate whether the synthesized audio accurately reflects the emotional transitions indicated in the corresponding caption. Use the following 5-point rating scale:
\begin{itemize}
    \item \textbf{5 points:} Completely consistent in both number and types of emotions.
    \item \textbf{4 points:} Consistent in the number of emotions, with partial consistency in emotion categories.
    \item \textbf{3 points:} Consistent in the number of emotions, but completely inconsistent in emotion categories.
    \item \textbf{2 points:} Inconsistent in the number of emotions, but the audio partially reflects the emotions indicated in the caption (e.g., audio conveys only \textit{angry}, while the caption indicates \textit{sad $\rightarrow$ angry}).
    \item \textbf{1 point:} Completely inconsistent in both number and categories of emotions.
\end{itemize}

\textbf{Acc$_{ETC}$ / Acc$_{ETT}$}. For each caption version provided, annotate the emotional transitions indicated by answering the following:
\begin{itemize}
    \item How many distinct emotions are reflected in the caption? (Enter numerical values such as 1, 2, 3, etc.)
    \item Specify the sequence of emotions clearly (e.g., \textit{angry $\rightarrow$ sad $\rightarrow$ happy}).
\end{itemize}

\subsection{A.4.2 Objective Metrics}
\textbf{EES$_{ET}^{1\rightarrow 3}$}. Since each test sample is accompanied by a manually transcribed transcript, we employ an ASR-based alignment strategy during the testing phase in place of the MTETR model for emotion diarization. Specifically, we first utilize a pre-trained ASR model (Whisper-large-v2)\footnote{\url{https://huggingface.co/openai/whisper-large-v2}} to extract character-level time boundaries. Based on the number of characters in each emotional segment, we estimate emotional timestamps and segment the synthesized audio accordingly. Each segment is then passed through the Emotion2vec model to extract its emotion embedding. We compute the similarity between each synthesized segment's embedding and the corresponding ground-truth embedding. The overall discourse-level consistency between the synthesized speech and the ground-truth emotional transition is quantified by multiplying the segment-wise similarities, resulting in a unified similarity score.

\begin{table*}
\centering
\small
\renewcommand{\arraystretch}{1.4}
\begin{tabular}{
    >{\raggedright\arraybackslash}m{0.18\linewidth} 
    >{\raggedright\arraybackslash}m{0.78\linewidth} 
}
\toprule
\textbf{Model} & \textbf{Caption} \\
\midrule
\textbf{SECap} & 
The voice is smooth and the speech rate is brisk, conveying a sense of pride and self-satisfaction. \\
\midrule
\textbf{SpeechCraft} & 
The young girl speaks in a low-pitched and pleasant voice, with moderate volume and slow pacing, subtly conveying a sense of pride and self-satisfaction. \\
\midrule
\textbf{EmoTransCap ($V_I$)} & 
The young girl begins with a low and soft tone, slowly expressing her inner sense of loss and melancholy, as if overshadowed by the changes of reality. Gradually, her tone becomes lighter, revealing a hint of relief and hope, as though she sees a turning point and regains her optimistic expectations for the future. \\
\midrule
\textbf{EmoTransCap ($V_D$)} & 
\textbf{[Global Description]}  

The speaker is a young girl whose voice exhibits a clear emotional transition. Initially, she conveys a sense of sadness and emotional downturn, characterized by a slow speech rate and low pitch. As the utterance progresses, her emotion gradually shifts toward a more positive and optimistic state. Her voice begins to reflect relief and joy, with a slightly elevated pitch and consistent pacing, illustrating a transition from sorrow to contentment.  

\textbf{[Partial Description]}  

Part 1 (00:00–00:07):  

The voice is low and slow, carrying a clear sense of sadness. The energy remains moderate, but the tone is somewhat suppressed, expressing emotions of loss and disappointment.  

Part 2 (00:07–00:18):  

The voice becomes lighter and more uplifting. Although the speech rate remains relatively slow, the pitch increases slightly, and the emotional tone reflects a sense of relief and joy, projecting a gentle optimism. \\
\bottomrule
\end{tabular}
\caption{Case Study: Comparison of generated captions from different models.}
\label{tab:case-study}
\end{table*}

\section{A.5 Case Study}
To further validate the model’s effectiveness in capturing discourse-level emotional dynamics, we conduct a case study. As shown in Table \ref{tab:case-study}, SECap provides a brief and coarse emotional label, which lacks specificity and contextual nuance. SpeechCraft offers slightly richer descriptions by incorporating acoustic features such as pitch and speech rate, but it still fails to reflect the full trajectory of emotional transitions. In contrast, EmoTransCap ($V_{I}$) captures a coherent emotional flow, highlighting the shift from melancholy to hope through subtle changes in tone and pacing. Notably, EmoTransCap ($V_{D}$) provides the most comprehensive representation by explicitly distinguishing between global and local emotional patterns. It accurately captures segment-level emotional variations and produces fine-grained captions that closely align with the underlying emotional progression of the speech. These findings highlight the superior performance of EmoTransCap, particularly its discourse-aware variant, in generating emotionally expressive and temporally aligned captions.
\end{document}